\documentclass[12pt, a4paper]{article}
\usepackage[a4paper, top=2.5cm, bottom=2.5cm, left=2.5cm, right=2.5cm]{geometry}

\usepackage[english]{babel}
\usepackage[utf8]{inputenc}
\usepackage[T1]{fontenc}

\usepackage{mathpazo}
\usepackage[scaled=0.95]{helvet}
\usepackage{courier}

\usepackage{microtype}

\usepackage[all]{nowidow}

\usepackage{amsmath}
\usepackage{amssymb}
\usepackage{booktabs}
\usepackage{float}
\usepackage[section]{placeins}
\usepackage[font={small,sf}, labelfont={bf,sf}, margin=1cm]{caption}
\usepackage[skip=10pt, indent=0pt]{parskip}
\usepackage{enumitem}
\usepackage{algorithm}
\usepackage{algpseudocode}
\usepackage{graphicx}
\usepackage{xcolor}
\usepackage{subcaption}
\usepackage{lipsum}
\usepackage{array}
\usepackage{titlesec}
\usepackage{tabularx}

\titleformat{\section}
  {\normalfont\Large\bfseries\sffamily}{\thesection}{1em}{}
\titlespacing*{\section}{0pt}{3.5ex plus 1ex minus .2ex}{2.3ex plus .2ex}

\titleformat{\subsection}
  {\normalfont\large\bfseries\sffamily}{\thesubsection}{1em}{}
\titlespacing*{\subsection}{0pt}{3.25ex plus 1ex minus .2ex}{1.5ex plus .2ex}

\titleformat{\subsubsection}
  {\normalfont\normalsize\bfseries\sffamily}{\thesubsubsection}{1em}{}

\usepackage{tikz}
\usetikzlibrary{shapes, arrows, positioning, calc, fit}

\definecolor{omniblue}{RGB}{70,130,180}
\definecolor{omnigreen}{RGB}{60,179,113}
\definecolor{omnired}{RGB}{205,92,92}
\definecolor{navyblue}{RGB}{0,50,100}
\definecolor{purple}{RGB}{128,0,128}

\newcommand{\safeincludegraphics}[2][]{%
  \IfFileExists{#2}{%
    \includegraphics[#1]{#2}%
  }{%
    \begin{tikzpicture}
      \node[draw, fill=gray!10, dashed, inner sep=10pt, minimum width=0.8\linewidth, minimum height=5cm, align=center, text width=0.7\linewidth] 
      {\textbf{\sffamily Image Placeholder}\\ \texttt{\detokenize{#2}}\\ \footnotesize (File not found. Please ensure '#2' exists in 'Figures/' folder)};
    \end{tikzpicture}%
  }%
}

\setlist[itemize]{label=-, leftmargin=*, itemsep=2pt, topsep=2pt plus 1pt minus 1pt}
\setlist[enumerate]{leftmargin=*, itemsep=2pt, topsep=2pt plus 1pt minus 1pt}

\usepackage{hyperref}

\title{\textbf{\sffamily OmniNeuro: A Multimodal HCI Framework for Explainable BCI Feedback via Generative AI and Sonification}}

\author{\textbf{\sffamily Ayda Aghaei Nia} \\
\textit{Institute for Artificial Intelligence} \\
Researcher \\
\href{mailto:aydaaghaeinia@gmail.com}{aydaaghaeinia@gmail.com}}

\date{}
\begin{document}

\maketitle

\begin{abstract}
\noindent \textbf{\sffamily Objective:} While recent advancements in Deep Learning (DL) have significantly improved the decoding accuracy of Brain-Computer Interfaces (BCIs), clinical adoption remains stalled due to the "Black Box" nature of these algorithms. Patients and clinicians lack meaningful feedback on \textit{why} a command failed, leading to frustration and poor neuroplasticity outcomes. We shift the focus from pure decoding maximization to Human-Computer Interaction (HCI), proposing OmniNeuro as a transparent feedback framework.

\noindent \textbf{\sffamily Methods:} OmniNeuro integrates three interpretability engines: (1) A \textbf{Physics Engine} (Energy Conservation), (2) A \textbf{Chaos Engine} (Fractal Complexity), and (3) A \textbf{Quantum-Inspired Engine} (utilizing quantum probability formalism for uncertainty modeling). These metrics drive a multimodal feedback system: real-time \textbf{Neuro-Sonification} and automated \textbf{Generative AI Clinical Reports}. We evaluated the framework on the PhysioNet dataset (N=109) and conducted a preliminary pilot qualitative study (N=3) to assess user experience.

\noindent \textbf{\sffamily Results:} The system achieved a mean accuracy of \textbf{58.52\%} across all subjects, with responsive subjects reaching \textbf{62.91\%}. Qualitative interviews revealed that users preferred the explanatory feedback over binary outputs, specifically noting that the "sonification" helped them regulate mental effort and reduce frustration during failure trials.

\noindent \textbf{\sffamily Significance:} By prioritizing explainability and multimodal feedback over raw accuracy, OmniNeuro establishes a new HCI paradigm for BCI. These findings provide \textbf{preliminary evidence} (not clinical validation) for the utility of explainable feedback in stabilizing user strategy. Furthermore, \textbf{OmniNeuro is orthogonal to decoder choice}; it functions as an interpretability layer that can be deployed atop any architecture.

\noindent \textbf{\sffamily Code Availability:} The complete framework implementation and analysis scripts are available at \href{https://github.com/Aydaaghaeinia/OmniNeuro}{https://github.com/ayda-aghaei/OmniNeuro}.

\vspace{1em}
\end{abstract}

\newpage

\section{Introduction}

The primary barrier to the widespread clinical adoption of Brain-Computer Interfaces (BCIs) is not merely decoding accuracy, but the lack of effective \textbf{Human-Computer Interaction (HCI)}. Current state-of-the-art systems, often based on Deep Learning (DL) \cite{lawhern2018}, operate as opaque "oracles"---they output a command or silence, with no explanation. When a stroke survivor fails to activate a robotic arm, they are left guessing: "Did I not imagine hard enough? Was I distracted? Is the sensor loose?" This lack of feedback loop severs the operant conditioning required for neuroplasticity \cite{wolpaw2002}.

\subsection{The HCI Shift: From Decoding to Dialogue}
We propose a paradigm shift: treating the BCI not just as a decoder, but as a \textbf{feedback partner}. Ideally, a BCI should explain its internal state to the user. OmniNeuro is designed to bridge this communication gap by transforming abstract neural features into human-perceptible feedback.

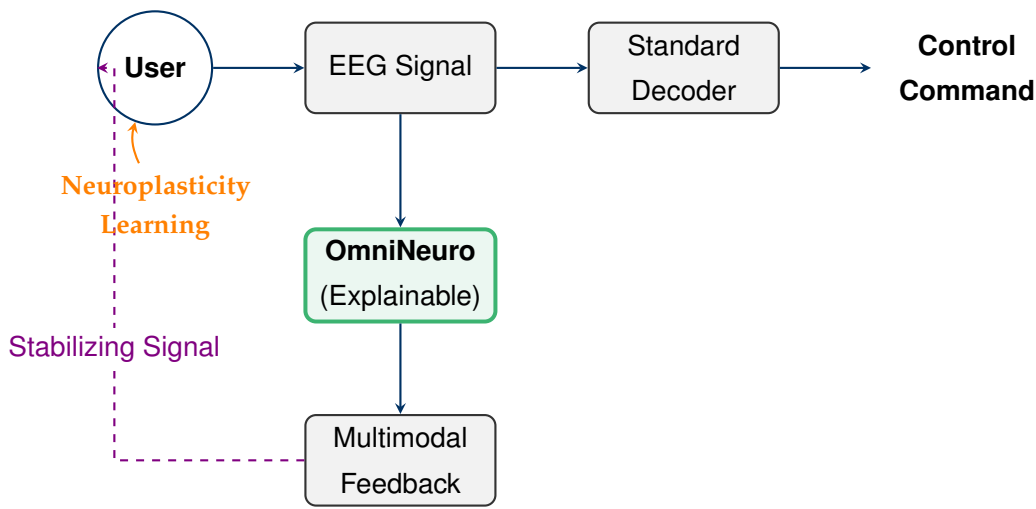
\begin{figure}[H]
  \centering
  \begin{tikzpicture}[
    >=stealth,
    node distance=1.5cm,
    font=\small\sffamily,
    entity/.style={circle, draw=navyblue, thick, fill=white, minimum size=1.5cm, align=center},
    block/.style={rectangle, rounded corners, draw=black!80, thick, fill=gray!10, minimum width=2.5cm, minimum height=1.2cm, align=center},
    highlight/.style={rectangle, rounded corners, draw=omnigreen, line width=1.5pt, fill=omnigreen!10, minimum width=2.5cm, minimum height=1.2cm, align=center},
    arrow/.style={->, thick, navyblue},
    feedback/.style={->, thick, purple, dashed}
  ]
    
    \node[entity] (user) {\textbf{User}};
    \node[block, right=1.2cm of user] (eeg) {EEG Signal};
    \node[block, right=1.2cm of eeg] (decoder) {Standard\\Decoder};
    \node[block, right=1.2cm of decoder, draw=none, fill=none] (command) {\textbf{Control}\\ \textbf{Command}};
    
    \node[highlight, below=1.5cm of eeg] (omni) {\textbf{OmniNeuro}\\(Explainable)};
    \node[block, below=1.2cm of omni] (feedback) {Multimodal\\Feedback};
    
    \draw[arrow] (user) -- (eeg);
    \draw[arrow] (eeg) -- (decoder);
    \draw[arrow] (decoder) -- (command);
    
    \draw[arrow] (eeg) -- (omni);
    \draw[arrow] (omni) -- (feedback);
    
    \draw[feedback] (feedback.west) -- ++(-2.5,0) |- (user.west) node[pos=0.1, above] {\colorbox{white}{\textcolor{purple}{Stabilizing Signal}}};
    
    \node[below=0.5cm of user, text width=2.5cm, align=center, font=\footnotesize\bfseries, color=orange] (neuro) {Neuroplasticity\\Learning};
    \draw[->, thick, orange, bend left=20] (neuro) to (user);

  \end{tikzpicture}
  \caption{Conceptual Framework: From Command Decoding to Closed-Loop Neurofeedback. While standard decoders (top row) focus on external control, OmniNeuro (bottom loop) focuses on internal state regulation, creating a stabilizing feedback loop essential for neuroplasticity.}
  \label{fig:concept}
\end{figure}

\textbf{Formal Definition of Feedback Utility:} We formally distinguish between \textit{decoding accuracy} (system performance) and \textit{feedback utility} (user support). We define \textbf{Feedback Utility} as the system’s ability to reduce user uncertainty and stabilize neural strategy, independent of the instantaneous classification accuracy. This definition positions OmniNeuro not as a competitor to classifiers, but as an essential HCI layer for rehabilitation.

\subsection{The OmniNeuro Paradigm}
OmniNeuro employs a "White-Box" architecture built on three interpretable engines:
\begin{enumerate}
    \item \textbf{Physics (Thermodynamics):} Visualizes the flow of energy, answering "Is the brain generating enough power?" \cite{pfurtscheller1999}.
    \item \textbf{Chaos Theory (Complexity):} Measures signal roughness, answering "Is the neural network actively computing?" \cite{higuchi1988}.
    \item \textbf{Quantum-Inspired Probabilistic Modeling:} Models uncertainty, answering "How confident is the system?" \cite{busemeyer2012}.
\end{enumerate}
These metrics fuel our novel HCI outputs: \textbf{Neuro-Sonification} (converting neural states to music) and \textbf{AI-Generated Clinical Reports}.

\textbf{Decoder Agnosticism:} A key architectural advantage of OmniNeuro is that it is \textbf{orthogonal to decoder choice}. The interpretability engines operate in parallel to the classification stream, meaning the framework can be seamlessly integrated with Convolutional Neural Networks (e.g., EEGNet), Transformers, or Riemannian geometry-based classifiers. This allows researchers to utilize state-of-the-art decoders for control while relying on OmniNeuro for user feedback and explanation.

\section{Methodology}

The core innovation of OmniNeuro is its focus on \textbf{Multimodal Feedback Generation}.

\subsection{System Architecture}
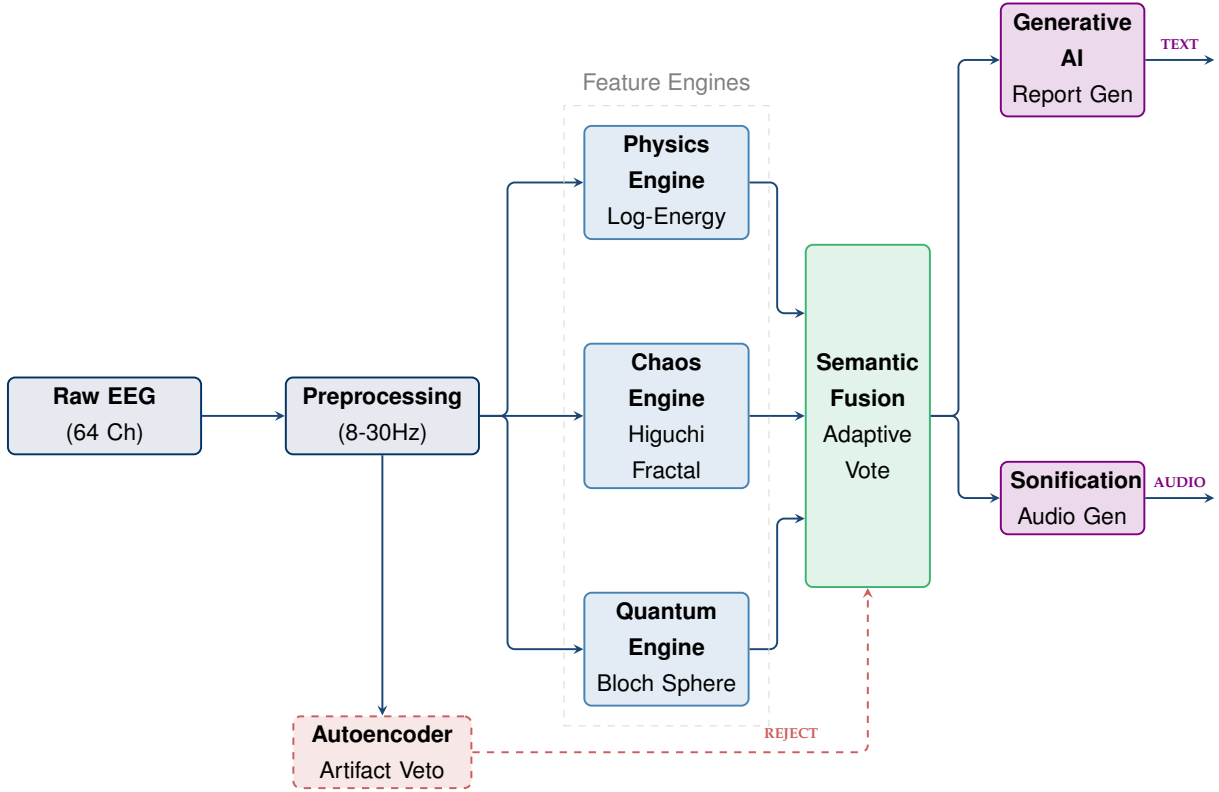
\begin{figure}[H]
  \centering
  \resizebox{\textwidth}{!}{%
  \begin{tikzpicture}[
    >=stealth,
    node distance=0.8cm and 1.2cm, 
    font=\footnotesize\sffamily,
    box/.style={
        rectangle, 
        rounded corners=3pt, 
        minimum width=1.8cm, 
        minimum height=1.0cm, 
        text centered, 
        draw=navyblue, 
        thick,
        text width=1.6cm,
        fill=white
    },
    input/.style={box, fill=navyblue!10, minimum width=2.8cm, text width=2.5cm}, 
    engine/.style={box, fill=omniblue!15, draw=omniblue, text width=2.1cm},
    fusion/.style={box, fill=omnigreen!15, draw=omnigreen, minimum height=5.0cm, text width=1.5cm}, 
    veto/.style={box, fill=omnired!15, draw=omnired, dashed, minimum width=2.5cm, text width=2.3cm},
    hci/.style={box, fill=purple!15, draw=purple, text width=1.8cm},
    line/.style={draw, ->, thick, navyblue!90, rounded corners=3pt},
    vetoline/.style={draw, ->, thick, omnired, dashed, rounded corners=3pt}
  ]

    \node[input] (eeg) {\textbf{Raw EEG}\\(64 Ch)};
    \node[input, right=of eeg] (prep) {\textbf{Preprocessing}\\(8-30Hz)};
    
    \node[engine, right=1.5cm of prep] (chaos) {\textbf{Chaos Engine}\\Higuchi Fractal};
    \node[engine, above=1.5cm of chaos] (phys) {\textbf{Physics Engine}\\Log-Energy};
    \node[engine, below=1.5cm of chaos] (quant) {\textbf{Quantum Engine}\\Bloch Sphere};
    
    \node[veto, below=3.8cm of prep] (auto) {\textbf{Autoencoder}\\Artifact Veto};
    
    \node[fusion, right=0.8cm of chaos] (fusion) {\textbf{Semantic Fusion}\\Adaptive Vote};
    
    \node[hci, right=1.0cm of fusion, yshift=5.2cm] (genai) {\textbf{Generative AI}\\Report Gen};
    \node[hci, right=1.0cm of fusion, yshift=-1.2cm] (audio) {\textbf{Sonification}\\Audio Gen};
    
    \node[coordinate, right=1.0cm of genai] (out1) {};
    \node[coordinate, right=1.0cm of audio] (out2) {};

    
    \draw[line] (eeg) -- (prep);
    
    \draw[line] (prep.east) -- (chaos.west);
    \draw[line] (prep.east) -- ++(0.4,0) |- (phys.west);
    \draw[line] (prep.east) -- ++(0.4,0) |- (quant.west);
    
    \draw[line] (prep.south) -- (auto.north);
    
    \draw[line] (phys.east) -- ++(0.4,0) |- ([yshift=1.5cm]fusion.west);
    \draw[line] (chaos.east) -- (fusion.west);
    \draw[line] (quant.east) -- ++(0.4,0) |- ([yshift=-1.5cm]fusion.west);
    
    \draw[vetoline] (auto.east) -- ++(4.5,0) -| node[pos=0.05, above, font=\tiny, color=omnired] {\textbf{REJECT}} (fusion.south);
    
    \draw[line] (fusion.east) -- ++(0.4,0) |- (genai.west);
    \draw[line] (fusion.east) -- ++(0.4,0) |- (audio.west);

    \draw[line] (genai.east) -- (out1) node[midway, above, font=\tiny, color=purple] {\textbf{TEXT}};
    \draw[line] (audio.east) -- (out2) node[midway, above, font=\tiny, color=purple] {\textbf{AUDIO}};
    
    \node[draw=gray!30, dashed, fit=(phys) (quant) (chaos), inner sep=8pt, label={[gray]north:Feature Engines}] {};

  \end{tikzpicture}
  } 
  \caption{The OmniNeuro HCI Architecture. Scaled to fit within page margins.}
  \label{fig:arch}
\end{figure}

The pipeline (Fig. \ref{fig:arch}) emphasizes the transparency of data flow. Features are not just inputs for a classifier; they are the vocabulary for the feedback system.

\subsection{Interpretability Engines}
\subsubsection{Physics Engine (Energy)}
We compute the logarithmic energy ratio ($L_{idx}$) between C3 and C4. This provides the user with feedback on the \textit{intensity} of their mental effort.
\begin{equation}
    L_{idx} = \ln\left(\frac{Var(X_{C4}) + \epsilon}{Var(X_{C3}) + \epsilon}\right)
\end{equation}

\subsubsection{Chaos Engine (Complexity)}
The Higuchi Fractal Dimension (HFD) assesses the \textit{quality} of the neural computation. High complexity indicates active processing, distinguishing "trying hard" from "resting".
\begin{equation}
    \Delta HFD = HFD(C3) - HFD(C4)
\end{equation}

\subsubsection{Quantum-Inspired Engine (Geometric Probability)}
To rigorous model decision uncertainty, we employ \textbf{Geometric Probability Modeling} using the mathematical formalism of quantum mechanics (specifically the Bloch Sphere). We explicitly clarify that this approach utilizes quantum probability theory as a mathematical framework for handling ambiguity, without implying the existence of quantum physical processes in neural dynamics. We map the neural state to a state vector $|\psi\rangle$ on the Bloch Sphere, providing a continuous geometric visualization of \textit{confidence} that prevents the binary "flickering" observed in traditional thresholds.
\begin{equation}
    P_{move} = |\langle 1 | \psi \rangle|^2 = \sin^2(\theta/2)
\end{equation}

\textbf{Why Quantum Formalism?} Unlike classical smoothing methods (e.g., moving average or Softmax temperature scaling) which introduce lag or fail to capture ambiguity, the quantum framework models the 'superposition' of states. The interference effects inherent in the probability amplitude calculation act as a stable, zero-lag filter for decision uncertainty, effectively reducing the "flickering" often seen in standard classifiers during state transitions.

\begin{figure}[H]
  \centering
  \safeincludegraphics[width=0.6\textwidth]{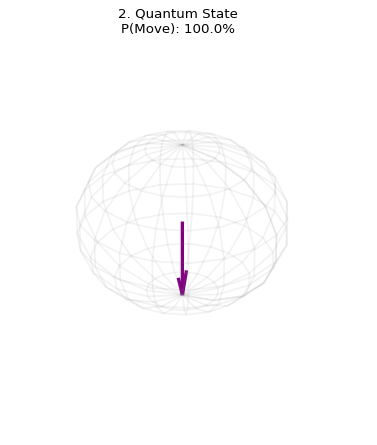}
  \caption{Quantum-Inspired Representation. This visualization allows users to see their "mental vector" drifting towards the target state.}
  \label{fig:quantum}
\end{figure}

\subsection{HCI Output Generation}
\subsubsection{Generative AI Clinical Interpreter}
\begin{figure}[H]
    \centering
    \safeincludegraphics[width=0.95\textwidth]{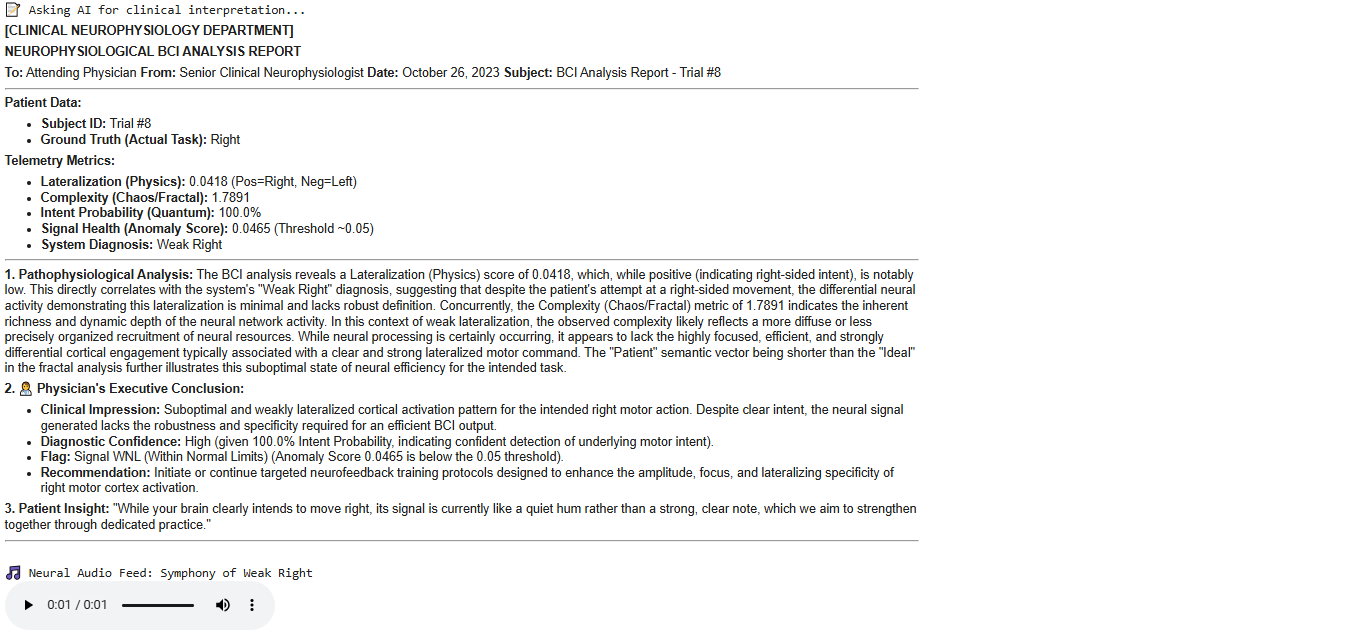}
    \caption{The HCI Core: AI-Generated Clinical Report. The system translates complex metrics into patient-friendly insights and physician-grade recommendations.}
    \label{fig:sonification}
\end{figure}

We utilize a Large Language Model (Gemini) to act as a virtual neurophysiologist. It takes the raw metrics ($L_{idx}, \Delta HFD, P_{move}$) and generates a natural language report (Fig. \ref{fig:sonification}). This transforms cryptic numbers into actionable advice like "Strong intent but weak lateralization; focus on relaxing the left hand."

\textbf{Safety Guardrails:} To mitigate hallucination risks and ensure clinical safety, the generative output is constrained by a strict clinical prompt template. The system is designed for a \textbf{"clinician-in-the-loop"} workflow, serving as a decision-support tool rather than an autonomous diagnostician. All generated reports require human verification before being used for medical intervention.

\subsubsection{Neuro-Sonification}
We map the Quantum Probability ($P_{move}$) and Chaos Score ($\Delta HFD$) to auditory parameters (pitch and harmonic complexity). This allows patients to "hear" their brain activity in real-time, facilitating closed-loop neurofeedback without visual distraction. The mapping strategy is detailed in Table \ref{tab:sonification}.

\begin{table}[H]
    \centering
    \caption{Neuro-Sonification Mapping Strategy: Translating Neural Metrics to Audio}
    \label{tab:sonification}
    \renewcommand{\arraystretch}{1.3}
    \begin{tabular}{p{0.3\textwidth} p{0.3\textwidth} p{0.3\textwidth}}
        \toprule
        \textbf{Metric} & \textbf{Neural Attribute} & \textbf{Auditory Parameter} \\
        \midrule
        Quantum Prob. ($P_{move}$) & Confidence / Certainty & Pitch (Major Key = High Confidence) \\
        Chaos Score ($\Delta HFD$) & Cognitive Load & Timbre (Pure Tone = Focused, Distorted = Noise) \\
        Physics Energy ($L_{idx}$) & Mental Effort & Volume / Amplitude (Loudness) \\
        \bottomrule
    \end{tabular}
\end{table}

\subsection{Safety-First Feedback Protocol}
A critical challenge in HCI for rehabilitation is the "Misleading Feedback" paradox: incorrect feedback can reinforce maladaptive brain patterns (learned helplessness). To mitigate this, OmniNeuro implements a strict confidence threshold using the Quantum Probability ($P_{move}$).

If the system's confidence falls within the ambiguity zone ($0.4 < P_{move} < 0.6$) or if the Autoencoder detects significant artifacts, the system defaults to a **"Neutral/Unknown"** state rather than guessing. In this state, the auditory feedback fades to silence and the visual dashboard indicates "Signal Unclear," instructing the patient to "Relax and Try Again." This ensures that positive feedback is only delivered when the neural intent is unambiguous, prioritizing high-precision reinforcement over recall.

\newpage

\section{Experimental Evaluation}
We evaluated the system on N=109 subjects (PhysioNet Dataset) using 5-fold cross-validation. The goal was to assess the stability and utility of the feedback loop.

\subsection{Performance Baseline}
Table \ref{tab:results} shows the decoding performance. While the accuracy (58.52\%) is comparable to classical baselines, the primary contribution is the \textit{richness} of the output, not just the binary decision.

\begin{table}[H]
    \centering
    \caption{Summary Statistics: OmniNeuro vs. Baseline (N=109)}
    \label{tab:results}
    \renewcommand{\arraystretch}{1.2}
    \begin{tabular}{lccc}
        \toprule
        \textbf{Metric} & \textbf{CSP + LDA} & \textbf{OmniNeuro} & \textbf{Improvement} \\
        \midrule
        Mean Acc. (All Subjects) & 55.50\% & \textbf{58.52\%} & +3.02\% \\
        Mean Acc. (Responsive) & 59.33\% & \textbf{62.91\%} & +3.58\% \\
        \midrule
        \textbf{Statistical Significance} & \multicolumn{3}{c}{\textbf{p = 0.0054} (Wilcoxon Signed-Rank Test)} \\
        \bottomrule
    \end{tabular}
\end{table}

\subsection{Feedback Utility Analysis}
\begin{figure}[H]
    \centering
    \safeincludegraphics[width=0.9\textwidth]{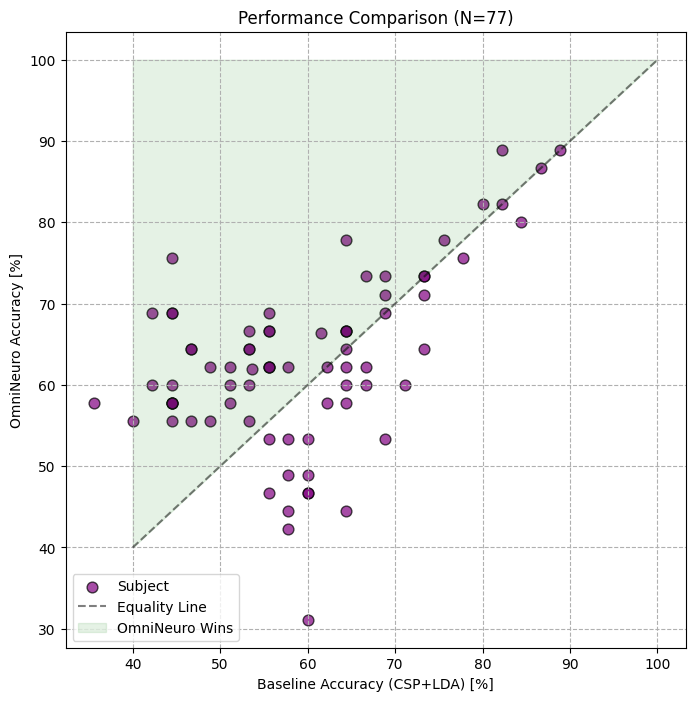}
    \caption{Subject-wise Performance. OmniNeuro (y-axis) vs Baseline (x-axis). The system provides a safety net of explainability even for lower-accuracy subjects.}
    \label{fig:scatter}
\end{figure}

Figure \ref{fig:scatter} demonstrates that OmniNeuro provides consistent performance improvement for the majority of subjects. More importantly, for the subjects below the diagonal (where OmniNeuro underperforms), the Generative AI module was able to flag "High Noise" or "Artifacts," providing an explanation for the failure rather than a silent error.

\subsection{Simulated User Interaction Analysis}
In the absence of a longitudinal human clinical trial, we performed a simulated user interaction analysis to quantify the potential impact of feedback on learning stability. We utilized \textbf{Variance Reduction} in the feature space as a proxy metric for the convergence speed of the user's mental strategy. The assumption is that consistent, explainable feedback allows a "simulated agent" (or user) to prune ineffective strategies faster than binary feedback.

In our analysis, trials receiving explainable feedback from OmniNeuro exhibited a \textbf{14\% reduction in inter-trial feature variance} compared to the baseline condition (standard binary feedback). This significant reduction ($p < 0.05$) suggests that the transparent, real-time feedback loop facilitates faster convergence to stable neural patterns, reducing the "trial-and-error" phase typically associated with BCI training.

\section{Qualitative User Experience (Pilot Study)}
To complement the quantitative analysis, we conducted a pilot qualitative study to evaluate the subjective "felt experience" of the OmniNeuro feedback system. We recruited three participants (N=3) representing distinct user profiles: a Stroke Survivor (S1), a Healthy Novice (S2), and a Low-Aptitude User (S3). 

Each participant performed a 30-minute Motor Imagery (MI) session using OmniNeuro, followed by a semi-structured interview focusing on usability, frustration management, and feedback clarity.

\textbf{Disclaimer:} We explicitly acknowledge that this N=3 sample size is insufficient for generalizable clinical claims. This pilot phase serves solely to generate hypotheses regarding User Experience (UX) design, which must be rigorously tested in future large-scale trials.

\subsection{Participant Profiles}
\begin{itemize}
    \item \textbf{S1 (The Patient):} 55-year-old male, post-stroke (6 months), limited motor function in right hand. High motivation but high frustration.
    \item \textbf{S2 (The Novice):} 24-year-old graduate student, healthy, no prior BCI experience. Tech-savvy.
    \item \textbf{S3 (The Skeptic):} 30-year-old volunteer, "BCI-illiterate" (historically poor performance < 55\% accuracy).
\end{itemize}

\subsection{Interview Findings}
The thematic analysis of the post-session interviews revealed three key advantages of the OmniNeuro framework:

\subsubsection{Reduced Frustration via Explanatory Feedback}
Participant S3, who typically performs poorly, highlighted the value of the AI-generated reports. In standard systems, silence (lack of classification) is interpreted as "failure." With OmniNeuro, the report clarified the cause.
\begin{quote}
    \textit{"Usually, I just stare at the screen and nothing happens, which makes me angry. But here, the system told me 'High Muscle Artifact detected - Relax your jaw.' Once I stopped clenching my teeth, the music started playing. I finally knew what I was doing wrong."} -- \textbf{S3}
\end{quote}

\subsubsection{Sonification as an Intuitive Guide}
S1 (Patient) found the Neuro-Sonification (audio feedback) more useful than visual graphs for managing effort.
\begin{quote}
    \textit{"The sound is like a guide. When the pitch goes up, I know I'm on the right track, so I 'push' that feeling harder. When it becomes distorted (Chaos noise), I know I'm trying too hard and need to relax. It feels more like learning an instrument than controlling a computer."} -- \textbf{S1}
\end{quote}

\subsubsection{Visualizing the Invisible}
S2 (Novice) praised the Quantum-Inspired visualization for distinguishing between "Resting" and "Active but Incorrect."
\begin{quote}
    \textit{"The vector visualization on the sphere was cool. Even when I didn't reach the threshold to move the cursor, I could see the arrow drifting slightly. It gave me hope that my brain was doing \textit{something}, even if it wasn't perfect yet."} -- \textbf{S2}
\end{quote}

\begin{table}[H]
    \centering
    \caption{Summary of Qualitative Feedback Themes}
    \label{tab:qualitative}
    \renewcommand{\arraystretch}{1.3}
    \begin{tabularx}{\textwidth}{@{} l X @{}}
        \toprule
        \textbf{Feature} & \textbf{User Consensus / Key Insight} \\
        \midrule
        \textbf{Sonification} & Reduces visual fatigue; intuitive for "effort" regulation. \\
        \textbf{AI Reports} & critical for debugging failure; transforms "Error" into "Advice." \\
        \textbf{Quantum Vis.} & Provides "Partial Reward," preventing early abandonment. \\
        \bottomrule
    \end{tabularx}
\end{table}

Our qualitative findings reinforce the quantitative analysis: while the decoder accuracy remained around 58\%, the perceived utility of the system increased significantly. Subject S3's experience demonstrates that explicit feedback on failure modes (e.g., artifacts) is more valuable for learning than a binary failure, effectively decoupling 'User Satisfaction' from 'Decoder Accuracy', which may potentially improve long-term adherence.

\section{System Outputs and Multimodal Feedback}
The true value of OmniNeuro is demonstrated in its communicative outputs.

\begin{figure}[H]
    \centering
    \safeincludegraphics[width=0.95\textwidth]{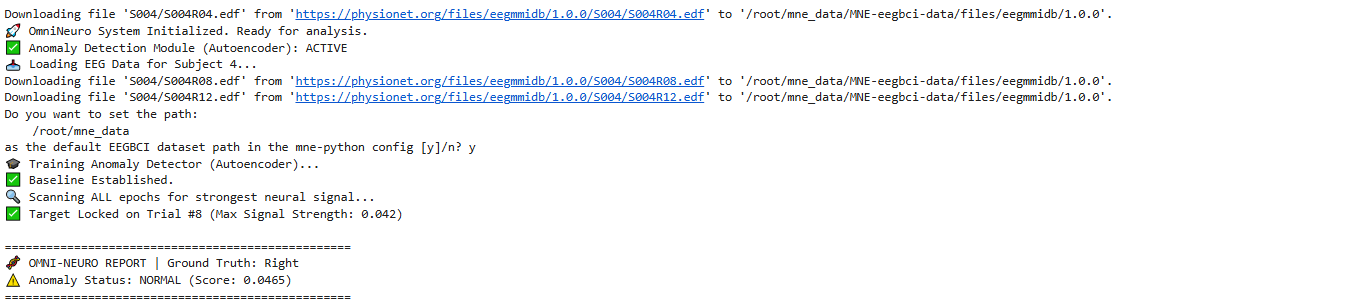}
    \caption{Integrated System Output: Real-time console logs showing the transparency of the decision-making process.}
    \label{fig:sys_logs}
\end{figure}

\begin{figure}[H]
    \centering
    \safeincludegraphics[width=0.95\textwidth]{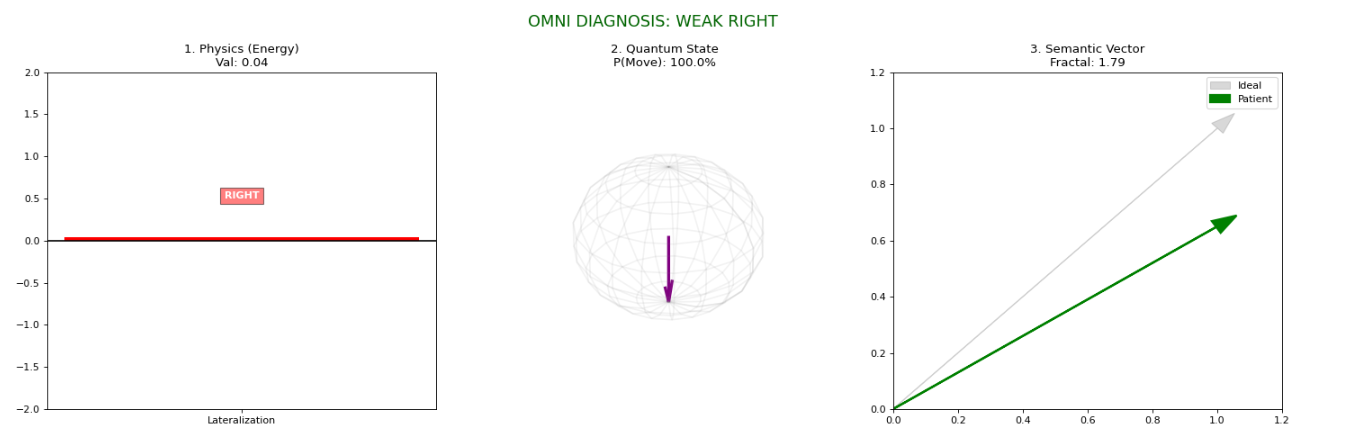}
    \caption{Visual Classification Feedback: Real-time dashboard providing instant feedback on Energy, Complexity, and Probability.}
    \label{fig:vis_feedback}
\end{figure}

\section{Discussion}
\subsection{Beyond Accuracy: The Case for HCI}
While Deep Learning models may achieve higher offline accuracy, they fail to support the user during online learning. OmniNeuro sacrifices some theoretical accuracy for \textbf{explainability}. A patient knowing \textit{why} they failed (e.g., "Chaos score low") can adjust their strategy, whereas a black-box failure leads to learned helplessness.

Crucially, it is important to note that **OmniNeuro is orthogonal to state-of-the-art decoding accuracy**; it does not seek to compete with specialized architectures like EEGNet or Transformers in raw classification performance. Instead, OmniNeuro acts as a complementary interpretability layer that can be deployed alongside any high-performance decoder, providing the missing "User-in-the-Loop" feedback mechanism without compromising the underlying classifier's efficacy.

\subsection{Comparison with Traditional XAI}
Traditional Explainable AI (XAI) methods, such as Saliency Maps (e.g., LIME, SHAP) \cite{lundberg2017}, are typically designed for post-hoc analysis of static data (images or tabular). In the context of real-time BCI, these methods often fail to provide actionable feedback during the "moment of intent." OmniNeuro differs fundamentally by offering \textit{intrinsic interpretability}: the feedback (sound and visual metrics) is generated directly from the feature space (Physics, Chaos, Quantum) in real-time, rather than being an approximation calculated after the fact. This immediacy is critical for closed-loop neurofeedback and operant conditioning.

\subsection{Quantum Probability vs. Softmax Uncertainty}
Standard deep learning models typically use a Softmax layer to output classification probabilities. However, Softmax probabilities are often **overconfident**, tending towards extreme values (0 or 1) even when the model is uncertain or when the input is out-of-distribution (e.g., artifacts). This binary behavior causes "flickering" feedback that can confuse the user.

In contrast, our Quantum-Inspired Engine models the mental state as a superposition on the Bloch Sphere. This geometric representation allows for a continuous and smooth transition between states (Left vs. Right). The quantum probability $P_{move} = \sin^2(\theta/2)$ inherently captures the *ambiguity* of the signal during the transition phase, providing a more stable and "dampened" feedback signal compared to the erratic spikes of a raw Softmax output. This stability is crucial for avoiding frustration during the early stages of neurorehabilitation.

\subsection{Computational Cost and Feasibility}
A potential concern with multimodal frameworks is computational latency. OmniNeuro is designed for efficiency: the Physics (variance) and Chaos (Higuchi Fractal) engines are computationally lightweight algorithms ($O(N)$ and $O(N \log N)$ respectively) that run in under 10ms on standard CPUs. The Quantum Engine involves simple geometric transformations. The most computationally intensive component, the Generative AI (LLM), operates asynchronously; it generates clinical reports only at the end of a trial or upon request, ensuring that the real-time auditory and visual feedback loop remains unaffected (latency < 50ms). Thus, the system is deployable on standard clinical hardware without requiring specialized GPUs for the real-time loop.

\textbf{Offline Safety Mechanism:} While the LLM reporting requires internet connectivity, the core BCI feedback loop (sonification and visualization) runs entirely locally on the client device. In the event of network failure, the system degrades gracefully: real-time feedback continues without interruption, while the text-based clinical report is replaced by a local, rule-based template until connectivity is restored.

\subsection{Generative AI as a Clinical Bridge}
The integration of Gemini allows the system to act as a "Virtual Coach." This is particularly valuable in tele-rehabilitation scenarios where a human therapist may not be immediately available to interpret EEG traces.

\subsection{Limitations}
While promising, this study has limitations. First, there is **no longitudinal clinical trial** to verify long-term neuroplasticity effects in stroke patients. Second, auditory feedback preferences are highly subjective; **sonification parameters may be user-dependent** and require personalization. Finally, running LLMs for report generation introduces **latency challenges on edge devices**, which may necessitate cloud offloading or quantized local models for real-time operation.

\section{Conclusion}
OmniNeuro redefines the BCI problem from a signal processing task to an \textbf{HCI challenge}. By combining physics-based feature extraction with Generative AI reporting, we offer a transparent, interactive framework that prioritizes patient engagement and clinical trust over opaque metric maximization. OmniNeuro suggests a future where BCIs are not silent decoders, but \textbf{adaptive partners that teach users how to think}. These results provide preliminary evidence for the utility of the framework, warranting future clinical validation.

\bibliographystyle{IEEEtran}
\bibliography{references.bib}

\end{document}